# Variable Stiffness Improves Safety and Performance in Soft Robotics

Mert Aydin
School of Mechanical, Materials, Mechatronic, and Biomedical Engineering,
University of Wollongong
Wollongong, NSW, Australia
ma731@uowmail.edu.au

Emre Sariyildiz
School of Mechanical, Materials, Mechatronic, and Biomedical Engineering,
University of Wollongong
Wollongong, NSW, Australia
emre@uow.edu.au

Charbel Dalely Tawk
School of Engineering,
Department of Industrial and Mechanical Engineering,
Lebanese American University
Byblos, Lebanon
charbel.tawk@lau.edu.lb

Rahim Mutlu
Faculty of Engineering and Information Sciences,
University of Wollongong in Dubai
Dubai, United Arab Emirates
rmutlu@uow.edu.au

Gursel Alici
School of Mechanical, Materials, Mechatronic, and Biomedical Engineering,
University of Wollongong
Wollongong, NSW, Australia
gursel@uow.edu.au

*Abstract*— **This paper proposes a new variable stiffness soft gripper that enables high-performance grasping tasks in industrial applications. The design of the proposed monolithic soft gripper includes a middle bellow and two side bellows (i.e., fingers). The positions of the fingers are regulated by adjusting the negative pressure in the middle bellow actuator via an on-off controller. The stiffness of the soft gripper is modulated by controlling the positive pressure in the fingers through the use of a proportional air-pressure regulator. It is experimentally shown that the proposed soft gripper can modulate its stiffness by 125% within 250ms. It is also shown that the variable stiffness soft gripper can help improve the safety and performance of grasping tasks in industrial applications.**

## I. INTRODUCTION

Compliant and soft robots have been widely used in various robotic systems (e.g., collaborative robots, exoskeletons, surgery robots and humanoids) to boost safety in physical robot-environment interaction [1–6]. Although their inherently compliant mechanical structures provide several benefits such as safety, and low-cost and high-fidelity force control, compliant and soft robots have certain challenges and fundamental limitations in motion control [5, 7 - 9]. For example, the elastic mechanical component limits the bandwidth of robotic systems, thus leading to a notable performance constraint in position control [8, 9]. As the stiffness of the elastic component decreases, the delay in the response time of the robotic system becomes larger. Moreover, the force range exerted by a compliant or soft robotic system is directly related to the stiffness of its elastic component. The softer elastic component a robotic system has, the lower force range it can exert in physical robot-environment interactions [5, 10]. Therefore, an elastic component with fixed stiffness leads to a significant trade-off between the safety and performance of compliant and soft robotic systems [5, 11].

To overcome such limitation in the soft robotics field and to make soft robots more useful for applications that require a range of force output [12] (i.e., from very small forces to interact with delicate and low-stiffness environments to high forces to interact with high-stiffness environments and to be able to lift moderate to high weights in case of a soft gripper [13] or a soft robotic hand [14]) soft robots are required to change their compliance actively and on demand (i.e., variable stiffness) [15]. For instance, a soft robotic gripper picking various objects [14] with different stiffness will be more efficient (i.e., apply the appropriate level of gripping force for each object) and more versatile (i.e., handle different loads and move them at different speeds or accelerations) with variable compliance [16].

Different approaches are considered to equip soft robots with variable stiffness capabilities [17]. Phase-changing materials [18] such as low point melting allows (LMPAs) have been employed to change the stiffness of a soft robot [19]. Another approach is to use polylactic acid (PLA) [20], shape-memory polymers (SMPs) [13, 21], and shape-memory alloys (SMAs) [22] which all soften when heated above their glass transition temperature. Similarly, wax [23] can be used to obtain variable stiffness soft structures for soft robotic applications. The main disadvantage of such approaches is their slow response and recovery since they are all thermally activated. In addition, electroactive polymers (EAPs) [24] and magneto-rheological fluids [25] have also been employed to achieve variable stiffness soft structures as well as bioinspired structures for variable stiffness integration in soft grippers [26, 27].

Other popular approaches are based on the jamming [28] of particles [29-31], layers [32], and fibers [33, 34] to achieve high stiffness levels and these are preferred the main actuation is achieved using a pneumatic power source (i.e., air compressor). Also, antagonistic structures have been employed for developing variable stiffness soft structures. This approach can be based on antagonistic pneumatic actuators [35, 36] and tendon-driven structures [37, 38]. To achieve fast response, multi-stage variable stiffness actuators have also been developed using bistable metal strips [39].

This work presents a fully-3D printed soft gripper with a single material that uses negative pressure for actuation and positive-pressure for varying the stiffness [40, 41]. The gripper is 3D printed using a fused deposition modeling (FDM) 3D printer and soft thermoplastic polyurethane. The soft gripper with variable stiffness capability was demonstrated using a robotic arm where it was used as an end-effector. The soft gripper proved that it can successfully grasp stably and firmly a water bottle at different picking and placing rates (i.e., speeds and accelerations) when the stiffness is augmented using positive pressure.

The remaining of the paper is organized as follows. Section II explains the design and manufacturing of the variable stiffness soft gripper. Section III describes the control system for grasping motion and stiffness modulation. Section IV evaluates the performance of the soft gripper experimentally. Sections V and VI provide a discussion and a conclusion.

## II. METHODS AND MATERIALS

### A. Design of the Variable Stiffness Soft Gripper

The soft robotic grippers are designed and modelled in Fusion 360 (Autodesk Inc.) based on soft bellows design as shown in Fig. 1. The soft robotic gripper operates pneumatically with positive as well as negative air pressure. The soft gripper has been designed by combining an actuation unit (i.e., negative pressure middle bellow) with soft fingers (i.e., positive pressure side bellows) which change their stiffness by pressure regulation. Thus, the soft gripper consists of two main components that are 3D printed simultaneously in one manufacturing step (i.e., monolithic soft gripper). The first component which is the middle shown in Fig. 1 is activated using negative pressure (i.e., vacuum) and it is the main driver or actuation unit. Once the middle bellow is activated, it contacts and pulls the two surrounding bellows inward since they are attached to it with soft ribs as shown in Fig. 1. The second component is the two bellow fingers that are activated using

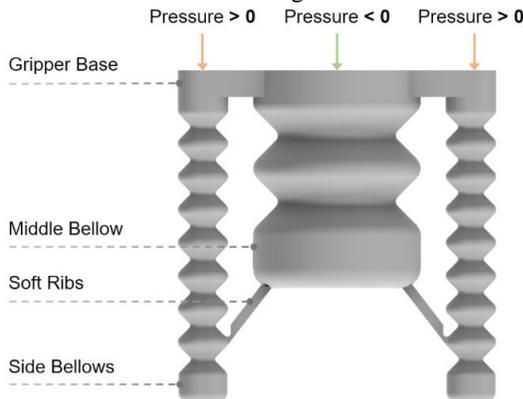
Figure 1: CAD design of the variable stiffness soft gripper.

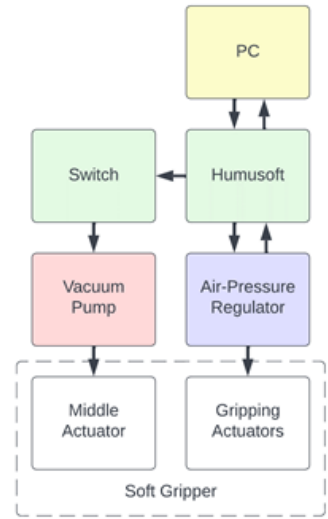
Figure 2: A flowchart that represents the control diagram of the proposed variable stiffness soft gripper.

positive pressure. These fingers generate a bending motion upon pressurization and consequently are pushed more inward when a positive pressure input is supplied. In addition, upon their activation with positive pressure their overall stiffness increases and consequently the stiffness of the gripper increases. Thus, these fingers act as the variable stiffness unit in the gripper. The reader is invited to refer to our previous study for further details on the design of the variable stiffness soft gripper [40].

### B. Materials and Additive Manufacturing Process

The soft grippers were printed using a low-cost and open-source fused deposition modelling 3D printer (FlashForge Inventor, FlashForge Corporation) that uses an off-the-shelf soft thermoplastic polyurethane (TPU) known commercially as NinjaFlex. NinjaFlex filament material provides stretchability and flexibility along its rigidity prior to printing which enables smoother printing process while allowing us to realize inflatable as well as deflate-able bellow-based soft robotic gripper with tuneable stiffness in this study. The printing parameters chosen to 3D print the soft grippers as airtight pneumatic soft structures are based on our previous studies which optimized key parameters for soft grippers and pneumatic actuators [40-43]. The grippers were 3D printed vertically along their height as illustrated in Fig. 1.

## III. CONTROL OF THE VARIABLE STIFFNESS SOFT GRIPPER AND ROBOTIC SYSTEM

The variable stiffness soft gripper is controlled using two independent controllers for tuning negative and positive pressures in position and stiffness modulations, respectively. A switch (i.e., an on-off controller) is used to control the positions of the fingers of the soft gripper. When the switch is turned on, the middle bellow actuator contracts and pulls the two side fingers inwards so that the gripper grasps the target object. In addition, a proportional air-pressure regulator is used to modulate the stiffness of the soft gripper by adjusting the positive air pressure in the fingers. As the positive pressure is increased, the soft gripper becomes stiffer. This allows us to perform safe and high-performance motion control tasks [5,

44]. The control structure of the variable soft gripper is illustrated in Fig. 2.

The grasping force of the variable soft gripper can be adjusted by increasing either the negative pressure in the middle bellow or the positive pressure in the side fingers. To increase the negative pressure, we can simply use a bellow with larger diameter. This enables us to carry heavier objects in industrial applications. The grasping force can also be adjusted by changing the stiffness of the gripper through positive pressure regulation. As the stiffness is increased through positive pressure regulation, the variable stiffness soft gripper can apply higher forces to objects. However, this is limited by the soft material characteristics and the design of the gripper. In this study, the pressure references were set between 50kPa to 350 kPa.

To evaluate the performance of the variable stiffness soft gripper in industrial applications, the gripper was mounted onto a collaborative robot. We used the Franka Emika Panda Collaborative Robot for the experimental demonstration in this study. It is worth noting that the proposed variable stiffness gripper can be easily mounted onto different industrial and collaborative robots. The Panda Robot was tasked to follow a trajectory using Franka Emika's high-level user interface. The programmed trajectories involve motions at various speeds and directions to test the soft gripper's capabilities when securely grasping and carrying an object. The experiments are detailed in Section IV.

## IV. EXPERIMENTS

This section verifies the performance of the proposed variable stiffness soft gripper experimentally. The experiments were conducted using a single-stage Rotary Vane vacuum pump to obtain a negative pressure source that controls the position of the gripper, a Chicago Air HUSH50 air compressor to generate positive pressure source for stiffness modulation, ME-system's K6D27 50/1Nm force sensor to estimate the force of the gripper, and Festo's VPPM-6L-10H proportional air-pressure regulator to control the stiffness of the gripper. A Real-time controller was implemented using MATLAB Simulink Desktop Real-Time and Humusoft MF644 Data Acquisition Card. The sampling time of the real-time control system was 1ms [45]. Pick and place tasks were performed using the Franka's Panda collaborative robot.

The first experiment is for stiffness modulation and it shows that the force applied by the gripper can be adjusted by modulating the stiffness of the griper via the proportional air-pressure regulator. In this experiment, the gripping force was measured by placing the force sensor at the tip point of the gripper while different positive pressures were applied by the regulator. This is illustrated in Fig. 3. This figure shows that the force exerted on the force sensor by the soft gripper increased ~100% when the positive pressure was increased from 150kPa to 350kPa. While the minimum gripping force the soft actuator could apply was ~7N when 50kPa positive pressure was applied to fingers, the maximum force was ~25N for 350kPa. The total gripping force increased more than three times when the positive pressure in fingers was changed from 50kPa to 350kPa. The soft gripper was capable of reaching the max force value within 250 ms as illustrated in Fig. 3.

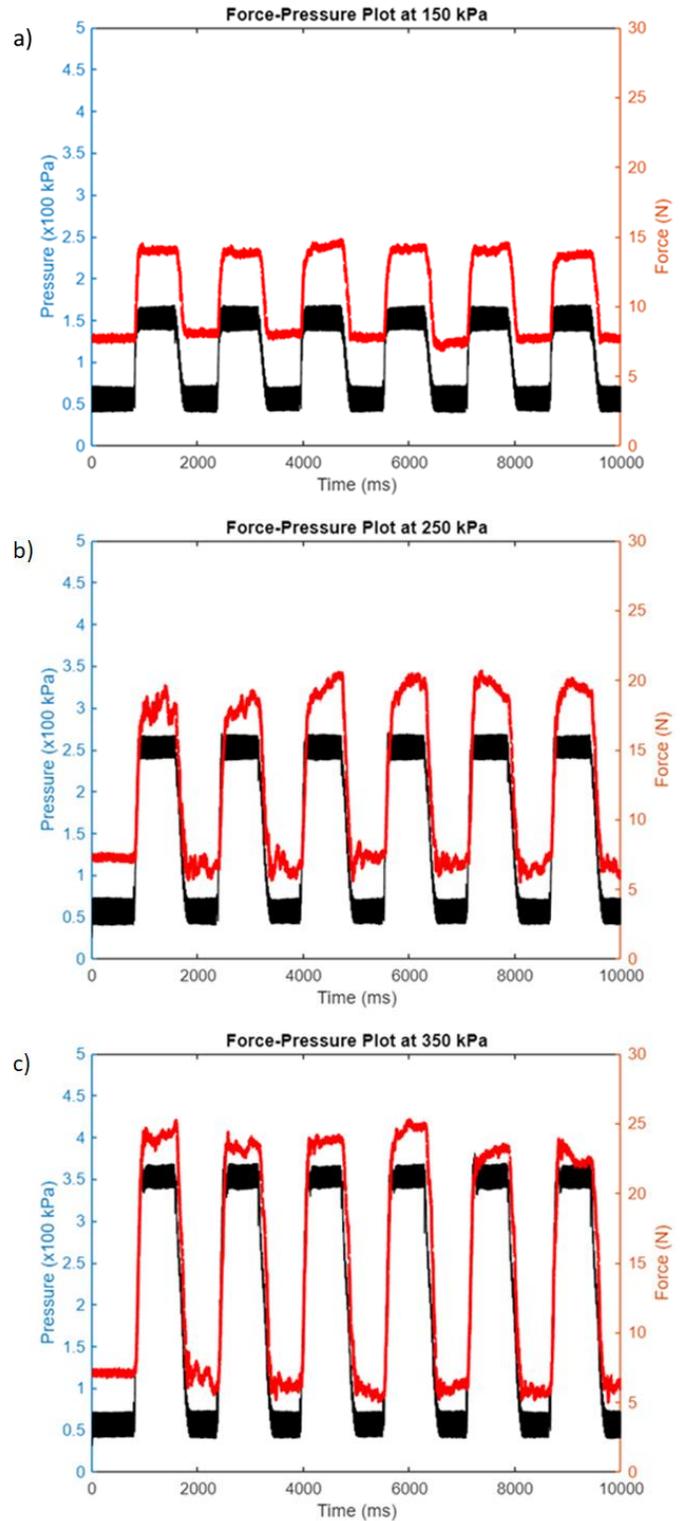

Figure 3: Stiffness modulation experiment 1. (a) Stiffness modulation experiment when 150kPa is applied to the soft gripper. (b) Stiffness modulation experiment when 250kPa is applied to the soft gripper. (c) Stiffness modulation experiment when 350kPa is applied to the soft gripper.

To estimate the stiffness variation, we applied 10N constant force to the gripper using a servo motor control system and measured the position deflection on the surface of the soft gripper while different positive pressures changing from 50kPa to 350kPa were applied through the proportional air pressure regulator. This experiment is illustrated in Fig. 4. While the softest mode was obtained as 0.2N/mm when the positive pressure in soft fingers was 50kPa, the stiffest mode of the soft gripper was 0.45N/mm for 350kPa. The proposed variable stiffness soft gripper allows us to modulate stiffness for 125%. It should be noted here that the stiffness of the soft gripper can be modulated from its minimum stiffness value 0.2N/mm to the maximum one 0.45N/mm within 250ms as shown in Fig. 3.

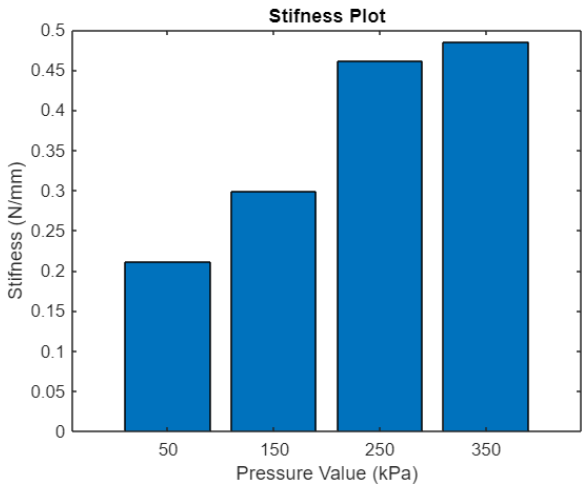

Figure 4: Stiffness modulation experiment 2.

To show how the variable stiffness property of the soft gripper enables us to improve the performance of practical engineering applications, two experiments were conducted where the softer and stiffer modes of the soft gripper provided different benefits when the gripper was mounted on a collaborative robotic arm to perform different industrial tasks. In the first experiment, the soft gripper performed a pick and place task for a fragile object (i.e., a disposable coffee cup) as shown in Fig. 5. The pressure of the soft gripper was set to 50kPa, 100kPa and 200kPa in Figs. 5a, 5b and 5c, respectively. As the stiffness of the soft gripper was increased, the force exerted on the disposable coffee cup became higher. While the soft gripper could safely grasp the disposable cup at 50kPa and 100kPa pressure values (see Figs. 5a and 5b), the cup was damaged when the pressure was set at 200kPa (see Fig. 5c). This figure clearly shows that even if a soft robot is used in grasping, the contact force between the robot and object should be properly adjusted to perform the pick and place task in a safe manner. The proposed variable stiffness soft gripper enables us to safely grasp fragile objects in its softer mode.

However, while the softer mode of the gripper improves safety in grasping fragile objects, it may significantly limit the performance of different pick and place tasks. An example of this limitation is illustrated in Fig. 6. In this experiment, a water bottle was grasped using the proposed variable stiffness gripper in the softer mode. When the pick and place tasks were performed at lower speeds (i.e., slower than 30 mm/s), the soft gripper could successfully complete the task. However, as the speed of the task was increased, (e.g., to higher than 45mm/s), the soft gripper failed in the softer mode for the task of placing the water bottle to the target location as shown in Fig. 6. To successfully complete this task at high speeds (i.e., at 45mm/s), stiffening mode was activated which significantly improved performing pick and place tasks of a water bottle as illustrated in Fig. 7.

## V. DISCUSSION

In this study, we have presented a single-material, monolithically 3D printed soft robotic gripper that has a variable stiffness property. The variable stiffness soft robotic gripper was fabricated using a low-cost FDM 3D printer along with a commercially available soft TPU. The soft robotic gripper operates by a combination of negative and positive air pressure bellows for actuation and stiffness modulation. Experimental tests were conducted to characterize the soft robotic gripper and to qualitatively demonstrate its gripping capability provided by stiffness modulation. While experimental characterization was conducted on bench setup focusing on relation between applied positive pressure and stiffness variation, and gripping force output measured at the tip of the soft robotic gripper fingers by gripping a multi-axis force sensor, qualitative tests were conducted by installing the gripper on a robotic manipulator as an end-effector to perform pick-and-place experiments.

Variable stiffness is a desirable property in robotic applications because it enables us not only to improve safety in robot-environment interactions but also to adapt the robotic systems (e.g., manipulators, grippers, and exoskeletons) to

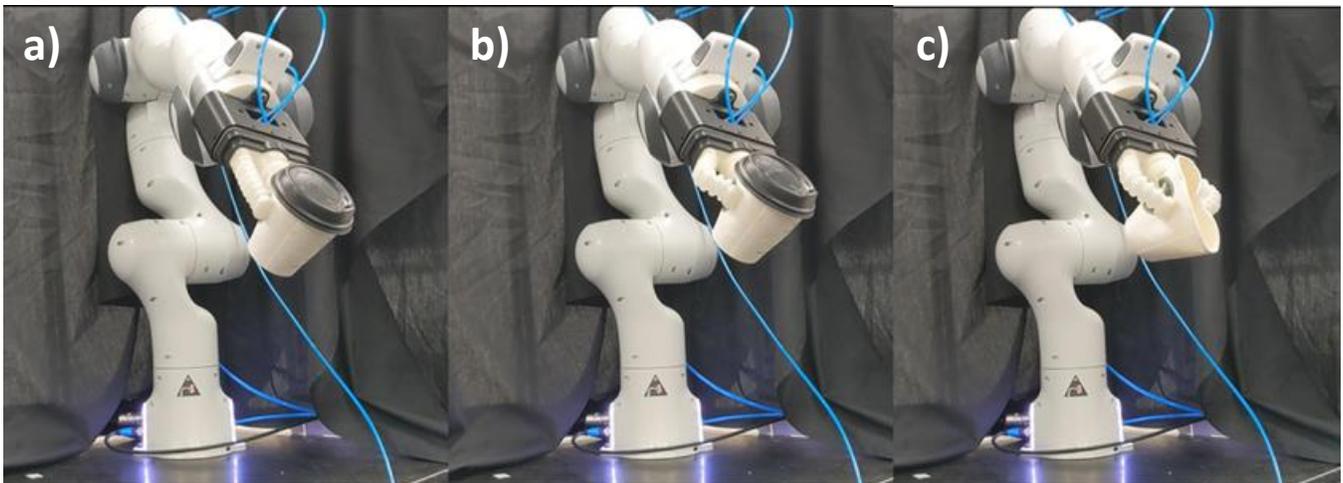

Figure 5: Grasping a disposable coffee cup experiment.

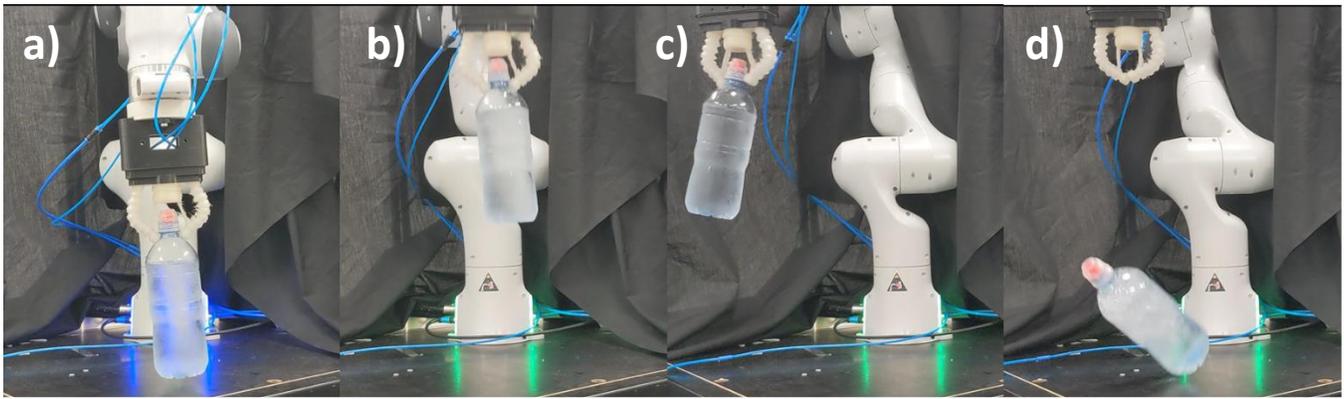
Figure 6: Pick and place experiment when the gripper works in the softer mode.

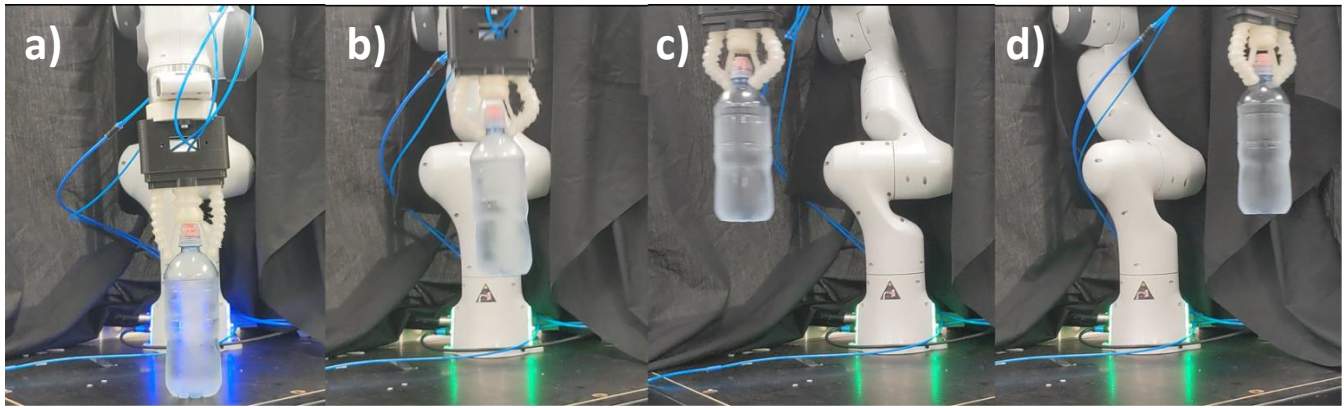
Figure 7: Pick and place experiment when the gripper works in the stiffer mode.

various different tasks in an effective manner [46]. For example, while a fragile object can be safely grasped using the softer mode of a variable stiffness robotic system, its stiffer mode simplifies the motion control problems by increasing the bandwidth of the closed-loop control systems and suppressing disturbances inherently [46]. This paper shows that the proposed monolithic soft gripper can provide the desirable variable stiffness property without the need of complex design and/or assembly procedures. Our soft gripper can change its stiffness from soft to its highest stiffness mode rapidly. Also, the results presented in Fig. 3 suggest that the stiffness modulation is repeatable even if rapidly changing stiffness is required for pick-and-place applications of multiple objects with different size, shape and stiffness. Thus our gripper can adapt to the designated tasks. In addition, the stiffness modulation results depicted in Fig.4 show that stiffness and input pressures relation up to 250kPa is quite linear, and over 250kPa it shows a saturation. Qualitative results demonstrating the pick-and-place capability of the soft robotic gripper in Figs. 5, 6 and 7 strongly advocate that stiffness modulation is significantly important not only for interacting with different objects but also for the speed at which a tasks is executed. As shown in Figs. 6 and 7, securely picking a filled water bottle by the gripper for higher speed movements (i.e., higher than or equal to 45mm/s) cannot be realized if the stiffness of the soft robotic gripper is not increased for the given design and experimental settings.

## VI. CONCLUSION

We have demonstrated a single-material, monolithically 3D-printed soft robotic gripper with intrinsic stiffness tunability in this paper. The property of variable stiffness is a common desire of roboticists which enables robotic systems to adapt to various tasks such as human-robot interaction and trajectory tracking in a safe and effective manner. This paper presents a series of experimental tests to quantitatively as well as qualitatively prove that the stiffness variation capability of the soft robotic gripper improves the safety and performance of soft grippers in industrial applications. Our future studies will include customization as well as scalability of such monolithic soft gripper design in further industrial settings.